\documentclass{article}

\usepackage{arxiv}

\usepackage[utf8]{inputenc} % allow utf-8 input
\usepackage[T1]{fontenc}    % use 8-bit T1 fonts
\usepackage{hyperref}       % hyperlinks
\usepackage{url}            % simple URL typesetting
\usepackage{booktabs}       % professional-quality tables
\usepackage{amsfonts}       % blackboard math symbols
\usepackage{nicefrac}       % compact symbols for 1/2, etc.
\usepackage{microtype}      % microtypography
\usepackage{lipsum}
\usepackage{graphicx}
\graphicspath{ {./images/} }

\usepackage{amsthm}
\theoremstyle{definition}
\newtheorem{definition}{Definition}

\newcommand{\model}{R\textsuperscript{6}}

\title{Open-Domain Question Answering \\with Pre-Constructed Question Spaces}

\author{
 Jinfeng Xiao \\
  University of Illinois at Urbana-Champaign\\
  \texttt{jxiao13@illinois.edu} \\
   \And
 Lidan Wang \\
  Adobe Inc. \\
  \texttt{lidwang@adobe.com} \\
  \And
 Franck Dernoncourt \\
  Adobe Inc. \\
  \texttt{dernonco@adobe.com} \\
  \And
 Trung Bui \\
  Adobe Inc. \\
  \texttt{bui@adobe.com} \\
  \And
 Tong Sun \\
  Adobe Inc. \\
  \texttt{tsun@adobe.com} \\
  \And
 Jiawei Han \\
  University of Illinois at Urbana-Champaign \\
  \texttt{hanj@illinois.edu} \\
}

\begin{document}
\maketitle
\begin{abstract}
Open-domain question answering aims at solving the task of locating the answers to user-generated questions in massive collections of documents. There are two families of solutions available: retriever-readers, and knowledge-graph-based approaches. A retriever-reader usually first uses information retrieval methods like TF-IDF to locate some documents or paragraphs that are likely to be relevant to the question, and then feeds the retrieved text to a neural network reader to extract the answer. Alternatively, knowledge graphs can be constructed from the corpus and be queried against to answer user questions. We propose a novel algorithm with a reader-retriever structure that differs from both families. Our reader-retriever first uses an offline reader to read the corpus and generate collections of all answerable questions associated with their answers, and then uses an online retriever to respond to user queries by searching the pre-constructed question spaces for answers that are most likely to be asked in the given way. We further combine retriever-reader and reader-retriever results into one single answer by examining the consistency between the two components.  We claim that our algorithm solves some bottlenecks in existing work, and demonstrate that it achieves superior accuracy on real-world datasets.
\end{abstract}

% keywords can be removed
\keywords{question answering, open-domain question answering, question space, reader-retriever}

\section{Introduction}

Open-domain question answering, abbreviated as \textit{OpenQA} in this paper, aims at enabling computers to answer user-submitted natural language questions based on a large collection of documents (a.k.a. a corpus). It has a wide range of applications including document understanding, user interaction, and decision making. It differs from the more ``standard'' question answering, abbreviated as \textit{QA} in this paper, where the task is to extract the answer from a given piece of text instead of a large corpus. The OpenQA task is ``open'' in the sense that the corpus can cover a massively wide range of topics and an ideal OpenQA engine should be open to answering any questions that are possibly answerable by some text in the corpus. While the research community and data-driven companies have long been pushing forward the cutting-edge research and industrialization of question answering methods, there remains large room for improvement under the open-domain setting.

\begin{figure}[t]
  \centering
  \includegraphics[width=0.7\textwidth]{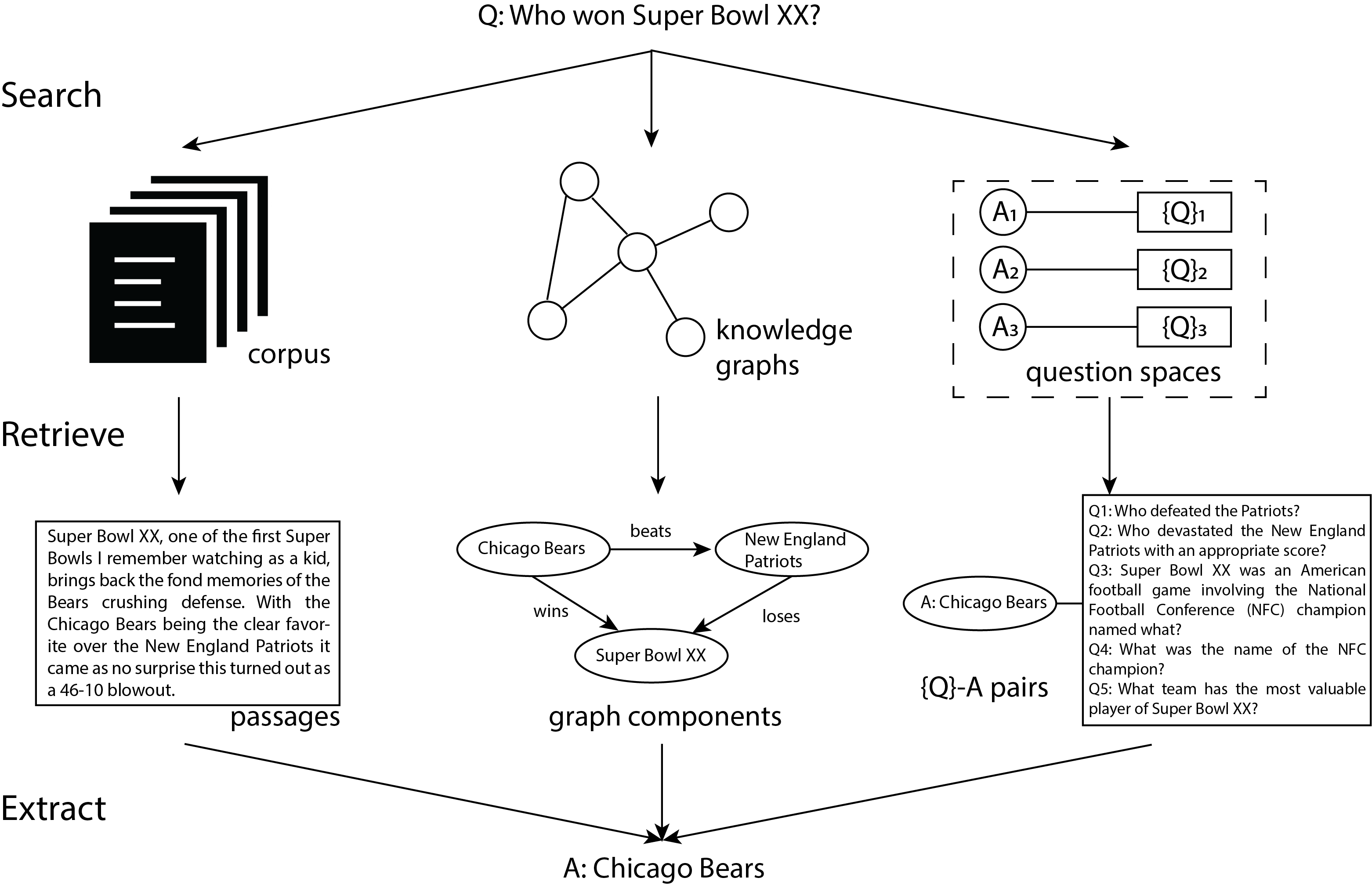}
  \caption{
  Retriever-readers (left), QA using KB (middle), and reader-retrievers (right).
  }
  \label{fig:1}
  \vspace{-0.4cm}
\end{figure}

There are two families of state-of-the-art OpenQA algorithms. One family, namely retriever-readers (Fig.\ \ref{fig:1}, left branch), first retrieves from the corpus some documents or paragraphs that are likely to be relevant to the question, and then uses neural networks to read the retrieved passages and locate the answer. Another line of work, namely question answering using knowledge bases (abbreviated as \textit{QA using KB} in this paper; Fig.\ \ref{fig:1}, middle branch), first constructs a knowledge base (KB) from the corpus, then rewrites the questions into some template format that is compatible with the KB schema, and finally queries the KB with the template for the answer. The knowledge base contains one or more knowledge graphs, which typically include typed entities and their relations. Either family of algorithms has some pros and cons to be discussed in later sections.

We propose a novel reader-retriever algorithm structure for OpenQA. The design, as shown with the right branch in Fig.\ \ref{fig:1}, differs from any existing work. First, we use deep neural networks to read the corpus offline, detect named entities, generate questions, and collect the results into two spaces of all questions that are answerable with the corpus. We use \textit{question spaces} to term the two spaces constructed in this way. When users enter queries online, a retriever compares user queries with the pre-constructed question spaces to retrieve the answers that are most likely to be asked in the given way. We then aggregate the resulted answers with answers generated by a traditional retriever-reader into one single answer to return to the user. The answer aggregation is based on the consistency between retriever-reader results and reader-retriever ones. Traditional retriever-readers are trusted more when there is some consistency, while our novel reader-retrievers are trusted more when the two sets of results do not agree at all. We call our final model \model~since it aggregates the results of one retriever-reader and two reader-retrievers. Experiments on public datasets show that the pre-constructed question spaces boost the performance for OpenQA, and \model~performs better than state-of-the-art methods by a large margin. Our main contributions include:

\begin{itemize}
    \item We propose a novel reader-retriever algorithm structure for OpenQA, where we first construct offline from the corpus two spaces of all answerable questions, and then retrieve online the answers that are most likely to be asked in the given way.
    \item We propose an answer aggregation mechanism to combine candidate answers returned by a retriever-reader and two different reader-retrievers into one final answer that is more reliable than any single candidate.
    \item We show with real-world datasets that the above procedure gives state-of-the-art OpenQA performance.
\end{itemize}

\section{Related Work} \label{sec:related}

\subsection{Open-Domain Question Answering} \label{sec:openqa}

Open-Domain Question Answering (OpenQA) aims at answering questions based on large collections of documents. It is harder than standard QA where the task is to find the answers from rather short text passages. While recent deep learning models have achieved over 85\% accuracy for QA \cite{zhang2020retrospective}, those models cannot be directly applied to OpenQA due to two constraints. First, it is computationally expensive and unnecessary to scan the entire corpus with a deep model for answering a specific question, whose answer is usually contained in just a few sentences. Second, the large body of irrelevant text will make the signal-to-noise ratio really small and thus result in bad accuracy, even if the entire corpus can be directly fed into a deep model. Thus, two families of algorithms have been specifically designed to solve OpenQA challenges: 1) retriever-readers, and 2) QA using KB.

\subsection{Retriever-Readers} \label{sec:retriever-readers}

It is natural to solve OpenQA by converting it to QA with information retrieval (IR) techniques, and that gives the popular retriever-reader algorithmic structure. One well-known example is DrQA \cite{chen2017reading}, which answers open-domain questions by first retrieving relevant Wikipedia pages with TF-IDF scores \cite{ramos2003using}, then reading the question and the retrieved page with recurrent neural networks (RNNs) \cite{hermann2015teaching,chen2016thorough}, and finally predicting the answer span by computing the similarity between the RNN-generated representations of the query and the retrieved paragraphs.

Unfortunately, although adding a retriever before a reader converts OpenQA to QA and thus makes existing QA methods readily applicable to OpenQA tasks, such a solution is far from perfect. All retriever-readers face an unavoidable trade-off between efficiency and accuracy. When the retriever module is computationally efficient (e.g. TF-IDF retrievers), the retrieved results will not be very reliable, and the performance of the subsequent reader module is constrained by the rather poor accuracy of the retriever \cite{htut2018training}. For example, while the exact-match accuracy of DrQA for standard QA is 69.5\% on SQuAD \cite{rajpurkar2016squad} and 49.4\% on Wikipedia, it dramatically decreases to 27.1\% under open-domain settings \cite{chen2017reading}. Such a big drop is largely due to errors made by the TF-IDF retriever used in DrQA. Another disadvantage of computationally efficient retrievers is that they lack trainable parameters for recovering from mistakes \cite{das2019multi}. On the other hand, there exist systems such as $R^3$ \cite{wang2018r} and DS-QA \cite{lin2018denoising} that have sophisticated retrievers jointly trained with the readers. Although those retrievers are more accurate, they are computationally expensive and thus not scalable to large corpora \cite{das2019multi}. To the best of our knowledge, there does not exist a scalable retriever that is accurate enough to bridge the gap between QA and OpenQA.

\subsection{QA Using KB} \label{sec:qa-using-kb}

There are solutions that solve OpenQA with knowledge bases (KB). QA using KB applications include Google Knowledge Graph and Bing Satori \cite{uyar2015evaluating}. Such approaches involve an offline knowledge graph construction module and an online graph query module. The graph construction module scans the entire corpus, detects entities, recognizes entity relations, and organizes those relations into a knowledge base that contains one or more knowledge graphs. Once a knowledge base is constructed, OpenQA tasks can then be converted to graph search tasks: finding the most likely answer is done by searching the most relevant node or edge in the graph. The search can be done in various ways, for example by template decomposition \cite{zheng2018question} or graph embedding \cite{huang2019knowledge}.  When a user submits a query question, the engine first rewrites the natural language query with some template format so that it can be used to search the KB, and extracts relevant entities or relations from the graph as the answer. A common query rewriting technique is to decompose natural language queries into events, each of which is described by a triplet $<$head entity, relation, tail entity$>$. For example, the query in Figure \ref{fig:1} can be rewritten into $<$\$person/org?, win, Super Bowl XX$>$, which means we want to query the KB for an entity with type ``person'' or ``organization'' (because the user asks ``who'') that has a relation ``win'' with the entity ``Super Bowl XX''.

There are a lot of challenges remaining for QA using KB. The common triplet decomposition of queries described above confines QA using KB to short factoid questions with easy syntax, and thus better query decomposition remains an active research topic. Another challenge is that QA using KB heavily relies on correct named entity recognition (NER) for the corpus and query, and errors in NER will propagate to later steps. Let's have another look at the example query in the previous paragraph. A correct response requires the correct recognition of the entity ``Super Bowl XX'' and the relation ``win'' in both the query and the corpus, as well as the entity ``Chicago Bears'' with the type ``organization'' in the corpus. If the NER module can only recognize ``Super Bowl'' but not ``Super Bowl XX'', which looks like just a small mistake, then the returned answer will likely be the winner for some other years which is completely wrong. There is another practical issue in applying QA using KB methods. Unlike in the retriever-reader world where some deep learning methods like BERT \cite{devlin2018bert}, XLNet \cite{yang2019xlnet} and their variants are clear winners, there are no widely accepted winners in the QA using KB family, and in many cases which QA using KB method works better depends on the actual corpus and task at hand. Despite that there is very little OpenQA literature that comprehensively compares QA using KB methods or compare them with retriever-readers, the community observes a recent trend that neural readers are dominating the leaderboards of public QA datasets but KB-based methods are not. Therefore, although QA using KB methods is a large family with great potential, we choose to focus on the comparison with retriever-readers when experimentally evaluating our proposed algorithm.

\subsection{Question Generation}

Another relevant line of work is question generation (QG). The task is to generate a question whose answer is a given text span in a given passage. Most recent approaches train encoder-decoders on public QA datasets like SQuAD so that the generated questions mimic the actual natural language questions \cite{kim2019improving,zhou2017neural,yuan2017machine}. While QG is a key component in our algorithm, the generated questions remain noisy, and how to generate questions with higher quality remains as interesting future work.

\section{Approach}\label{sec:approach}

To solve the challenges faced by existing work as discussed in Section \ref{sec:related}, we propose a reader-retriever model structure and an answer aggregation mechanism. Before making any online responses, the entire corpus is scanned offline by a deep learning QG reader to generate all answerable questions and collect them into two question spaces. After that, our algorithm responds to online queries by retrieving from the question spaces the answers that are most likely to be asked in the given way. Because the reader is now executed before the retriever, we believe we have found a path that avoids the efficiency-accuracy trade-off faced by state-of-the-art retriever-readers as discussed in Section \ref{sec:retriever-readers}. Compared to QA using KB methods, our question spaces are more flexible and friendly to natural language queries than knowledge graphs are, and our method is less prone to errors in NER because it requires correct recognition of the answer entity only instead of all entities in an event triplet.

\subsection{Question Spaces}\label{sec::question-spaces}
\begin{definition}\label{def:question-spaces}
A \textit{question space} is a bipartite graph with two disjoint and independent node sets $A$ and $Q$ representing the answers and associated questions. There are two types of question spaces: QA Spaces and \{Q\}A (read as Q-set-A) Spaces. In a \textit{QA Space}, each element $a_{i,j}$ of $A$ represents the $j$th occurrence in the corpus of the $i$th distinct named entity, and each element $q_{i,j}$ of $Q$ is a question generated from the context of $a_{i,j}$ with $a_i$ as its answer. For every $i$ and $j$, $a_{i,j}$ and $q_{i,j}$ form a \textit{QA pair} and are connected in the graph. In a \textit{\{Q\}A Space}, each element $a_i$ of $A$ represents the $i$th distinct named entity, and each element ${q}_i$ of $Q$ is a collection of the $q_{i,j}$'s for all $j$ in the QA Space. For every $i$,  $a_i$ and ${q}_i$ form a \textit{\{Q\}A pair} and are connected in the graph. In short, a QA space contains pairs of answers and generated questions, while a \{Q\}A space contains pairs of distinct answers and collections of all generated questions with that answer.
\end{definition}

For example, given the five questions in the right branch of Fig. \ref{fig:1} whose answer is ``Chicago Bears'', the QA Space will have five QA pairs: \{$a_{1,1}$ = ``Chicago Bears'', $q_{1,1}$ = ``Who defeated the Patriots?''\}, ..., \{$a_{1,5}$ = ``Chicago Bears'', $q_{1,5}$ = ``What team has the most valuable player of Super Bowl XX?''\}, and the \{Q\}A space will have one \{Q\}A pair: \{$a_1$ = ``Chicago Bears'', ${q}_1$ = \{``Who defeated the Patriots?'', ..., ``What team has the most valuable player of Super Bowl XX?''\}\}.

\subsection{Algorithm} \label{sec:detailed-structure}

% Figure: Detailed model structure
\begin{figure}[t]
  \centering
  \includegraphics[width=0.7\textwidth]{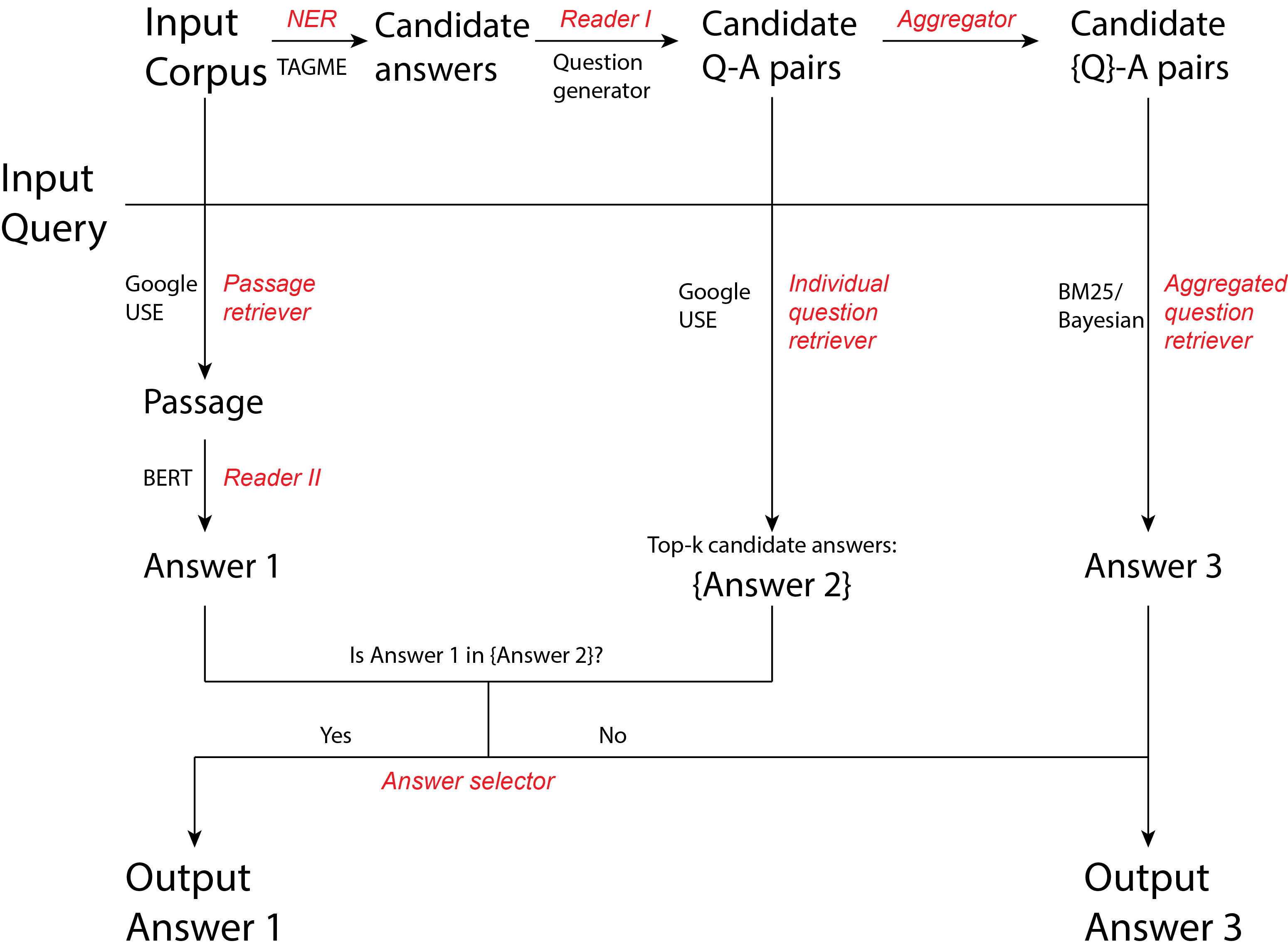}
  \caption{
  Detailed structure of the proposed method.
  }
  \label{fig:model-structure}
  \vspace{-0cm}
\end{figure}

A detailed illustration of our algorithm is given in Figure \ref{fig:model-structure}. The components above the grey dashed line are offline. They construct the QA Space and the\{Q\}A Space as defined in Definition \ref{def:question-spaces}. The modules below the grey dashed line are all executed online.

\subsubsection{NER, QG Reader and Question Aggregator} \label{sec:offline_reader}
Given a corpus, a named entity recognition tool called TAGME \cite{ferragina2010tagme,ferragina2011fast} is applied to detect named entities from the corpus and link the entities to Wikipedia titles. Those entities form the set of candidate answers $A$ in Definition \ref{def:question-spaces}. Then a question-generating reader is applied to the set of candidate answers to generate a question for each answer based on the local context around the answer phrases. The QG reader features an encoder-decoder model structure with a question-answering reward and a question fluency reward tuned with policy gradient optimization \cite{yuan2017machine,hosking2019evaluating}. This finishes the construction of the QA Space. We then use a question aggregator to build the \{Q\}A Space by putting together all the questions with the same answer.

\subsubsection{Passage Retriever and QA Reader}
Given a query, the passage retriever uses the dot product of the query embedding and passage embedding vectors generated by Google Universal Sentence Encoder (Google USE) \cite{cer2018universal} to retrieve from the corpus a passage that is semantically most similar to the query. We then use BERT \cite{devlin2018bert}, fine-tuned on SQuAD, to read the retrieved passage, predict the answer, and record the predicted answer as \textit{Answer 1}. We call this reader the QA Reader because other reader models like XLNet \cite{yang2019xlnet} that are suitable for standard QA tasks can also be used here. The pipeline that goes from Input Corpus to Passage and then Answer 1 is a valid retriever-reader workflow, and we denote this workflow as \textbf{Retriever-Reader-BERT-Large} or \textbf{Retriever-Reader-BERT-Base}, depending on which BERT model is used.

\subsubsection{Individual Question Retriever}
Given a query, the individual question retriever uses Google USE to retrieve from the QA space $k$ questions that are semantically most similar to the query. We record the collection of answers associated with the $k$ retrieved questions as \textit{\{Answer 2\}}. Note that \{Answer 2\} is an ordered list instead of a set, in the sense that its elements are ordered and duplicates are allowed. Although not illustrated Figure \ref{fig:model-structure}, a majority vote (where ties are resolved by average orders) over \{Answer 2\} can produce a single answer. We denote that single answer as \textit{Voted Answer 2}. 
%A valid reader-retriever workflow thus consists of the input corpus, candidate answer and QA Space generation (Section~\ref{sec:offline_reader}), and answer post-processing/voting.
Then the pipeline in Figure\ \ref{fig:model-structure} that goes from Input Corpus to Candidate Answers, QA Space, \{Answer 2\}, and finally Voted Answer 2 is a valid reader-retriever workflow. 
We denote this workflow as \textbf{Reader-Retriever-QA-Space}.

\subsubsection{Aggregated Question Retriever}
Given a query, the aggregated question retriever uses the BM25 score \cite{robertson2009probabilistic} to retrieve from the \{Q\}A space the answer whose associated set of questions is most similar to the given query. Note that if a distinct answer candidate entity $a_i$ occurs many times across the corpus, then its corresponding question set $q_i$ can be very big. We query the \{Q\}A Space by treating each $q_i$ as a single document which contains $q_{i,j}$ for all $j$ as sentences. In practice, we observe that BM25 works better for long documents and Google USE works better for short passages. That is why we use BM25 as the aggregated question retriever but use Google USE for the passage retriever and the individual question retriever. We record the answer $a_i$ associated to the top-ranked question set $q_i$ as \textit{Answer 3}. The pipeline that goes from Input Corpus to Candidate Answers, QA Space, \{Q\}A Space and finally Answer 3 is a valid reader-retriever workflow. We denote this workflow as \textbf{Reader-Retriever-\{Q\}A-Space}.

\subsubsection{Answer Aggregator}
Now that we have Answer 1, \{Answer 2\}, and Answer 3, the last step is to aggregate them into one single answer to return to the user. Our answer aggregation works as follows: if Answer 1 appears in the set \{Answer 2\}, then accept Answer 1 and return it; otherwise reject Answer 1 and return Answer 3. In other words, the answer aggregator checks the consistency between the retriever-reader results and the reader-retriever results, trust the retriever-reader if they agree to some extent, and trust the reader-retriever instead if the results do not agree at all. We name the complete workflow depicted in Figure \ref{fig:model-structure} as \textbf{\model}, meaning that the final result is contributed by a retriever-reader and two reader-retrievers. This is also the final model we deliver to the users.

\section{Experiments}

In this section, we evaluate the OpenQA performance of our proposed method \model~on a public dataset TriviaQA. On one hand, we compare its OpenQA accuracy against two state-of-the-art retriever-reader models, namely BERT and DrQA. On the other hand, we also examine the performance of the individual reader-retriever components inside \model~to get a sense of how well this novel structure is working and how much the answer aggregation process contribute. Finally, we give a qualitative discussion on the quality of the constructed question spaces. 

%\subsection{Goals and Design Principles}\label{sec::principles}
%The goal of our experiments is to answer the following questions: Can our proposed method \model~reach state-of-the-art OpenQA performance? If yes, how much do our newly proposed reader-retriever structure and answer aggregation mechanism contribute to the performance? %What are the aspects that deserve attention for future improvement of the method?

%In other words, we are mainly interested in examining the effectiveness and bottlenecks of the overall model structure as highlighted in red in Figure \ref{fig:model-structure} instead of individual modules highlighted in black. Therefore, we would like to keep away from distracting questions about individual modules such as ``which NER method works best for answer candidate detection'' or ``what is the best hyperparameter setting for a specific neural reader'' that are important in practice and time-consuming but do not really contribute much to answering the questions we care the most.

%Therefore, when designing the experiments, we keep the following principles in mind:

%\begin{itemize}[label={}]
%    \item Design Principle \#1. Given a task, consider approaches from the least controversial candidates.
%    \item Design Principle \#2. Regarding neural networks, use pre-trained models or follow author-suggested training settings whenever possible.
%\end{itemize}

%The practical implications of those principles will become more apparent to readers in the following sections.

\subsection{Data}
\subsubsection{TriviaQA}
TriviaQA \cite{joshi2017triviaqa} is a large-scale public dataset for reading comprehension and QA. It has a reading comprehension version and an unfiltered version, and in this paper we only discuss the reading comprehension version. It contains over 95,000 question-answer pairs supported by over 650,000 evidence documents, which means on average each question-answer pair is supported by seven documents. The documents are either Wikipedia articles or webpages. Each question-answer pair is associated with the list of documents that can be used to answer the given question.  This dataset can be used for OpenQA by simply putting all documents into one single corpus and leaving the questions as they are. We denote the open-domain version of the TriviaQA dataset as \textbf{TriviaQA-Open}.

\subsubsection{SQuAD}
The Stanford Question Answering Dataset (SQuAD) \cite{rajpurkar2016squad} is a very well-known public dataset for machine comprehension of text. It has two versions, and in this paper we only discuss the 1.1 version. It contains 107,785 question-answer pairs from 536 Wikipedia articles. Each question-answer pair is associated with a paragraph from which a human is able to find out the correct answer to the question. Each answer is a text span inside the given paragraph. The quality of this dataset is ensured by crowdsourcing, which means all questions are generated by humans and answered by humans. The scale and quality of SQuAD have made it one of the most popular datasets for training and evaluating QA and QG algorithms.

\subsubsection{Train/Test Split}
In our experiments, SQuAD serves as the training dataset for the QG reader and the reader module in DrQA, as well as the tuning dataset for BERT. On the other hand, TriviaQA-Open is used for evaluating all OpenQA methods. In other words, neural readers involved in this study are all trained or tuned with SQuAD, and they cannot see TriviaQA-Open data until evaluation.

Our choice of training and testing datasets is mainly based on two facts. 1) Using TriviaQA for testing makes NER an easier task. The TriviaQA authors report that about 93\% of the true answers in TriviaQA are Wikipedia titles. Meanwhile, TAGME \cite{ferragina2010tagme,ferragina2011fast} is specialized in annotating text with Wikipedia titles. Because we want the set of detected entities to overlap the set of true answers as much as possible, TAGME is a perfect NER tool for TriviaQA. NER on other datasets is harder, and how to do better NER is out of the scope of this study. 2) Using SQuAD for training enables us to utilize pre-trained models or author-suggested training settings to the greatest extent, so that we can make sure we correctly reproduce others' work and do not put their models into disadvantages when comparing them with ours. More details on this are available in Section \ref{sec:reproducibility}.

There may exist concerns about using different datasets for training and testing. We think that because our goal of the experiments is to check the effectiveness of our proposed methods and compare it with others, as long as all the methods are evaluated fairly under the same setting, we can achieve our goal. Such experimental settings are also used in the DrQA paper \cite{chen2017reading}. Although not critical to this study, using different datasets for training and testing has one additional benefit that it mimics the realistic application scenario where algorithms seldom know exactly how users actually behave at its training time. On the other hand, if the goal were to strive for the best possible performance on some dataset, training on another portion of the same dataset whenever possible would be the correct practice.
 
\subsection{Models}
We evaluate six different OpenQA methods with the exact match accuracy in the predicted answers on TriviaQA-Open. Five of them are introduced in Section \ref{sec:approach}, and the other is DrQA as introduced in Section \ref{sec:related}. Here we summarize the basic structure of all six methods in Table \ref{tab:models_structures}.

\begin{table}[ht]
  \centering
  \caption{Model structures. Arrows show orders of modules.}
  \label{tab:models_structures}
  \begin{tabular}{cc}
    \toprule
    Model & Description\\
    \midrule
    \model~& One retriever-reader plus two reader-retrievers\\
    DrQA & TF-IDF retriever $\rightarrow$ RNN reader\\
    Retriever-Reader-BERT-Large & Google USE retriever $\rightarrow$ BERT-large reader\\
    Retriever-Reader-BERT-Base & Google USE retriever $\rightarrow$ BERT-base reader\\
    Reader-Retriever-QA-Space & QG reader $\rightarrow$ Google USE retriever\\
    Reader-Retriever-\{Q\}A-Space & QG reader $\rightarrow$ Question aggregator $\rightarrow$ BM25 retriever\\
  \bottomrule
\end{tabular}
\end{table}

\subsection{Reproducibility Notes} \label{sec:reproducibility}
This section aims at providing as many details as possible that are needed to reproduce our results. All experiments are run on an Ubuntu 16.04 machine with eight GeForce GTX 1080 GPUs (CUDA version 10.1) and 24 CPUs. The entity score threshold for TAGME is set at 0.2 by tuning that value and manually inspecting the NER quality for 20 documents sampled from TriviaQA. The $k$ value for the individual question retriever that generates \{Answer 2\} is set to 10. For TriviaQA, we treat each paragraph with at least 50 characters as a passage, and drop paragraphs shorter than that. BERT is downloaded from the pytorch-transformers GitHub repository\footnote{https://github.com/huggingface/transformers} and fine-tuned on SQuAD following the documentation. The QG reader is obtained from the question-generation GitHub repository\footnote{https://github.com/bloomsburyai/question-generation} and trained on SQuAD with default settings. DrQA codes are downloaded from its GitHub repository\footnote{https://github.com/facebookresearch/DrQA}, the model trained by the authors on SQuAD is obtained as instructed, and the hyperparameter n-docs is set to 1 at prediction time for fair comparisons with \model. The BM25 retriever is implemented with ElasticSearch. The Google USE retrievers are implemented by first using BM25 to get the top ten thousand passages and then using dot products between Google USE embedding vectors to rank them. Google USE embeddings are performed by calling it from TensorFlow\footnote{https://tfhub.dev/google/universal-sentence-encoder/2}.

\subsection{Overall Test Accuracy}
Table \ref{tab:accuracy_overall} reports the overall test accuracy on TriviaQA-Open of our proposed method \model, three state-of-the-art methods (DrQA, Retriever-Reader-BERT-Large, and Retriever-Reader-BERT-Base), and the two novel workflows we introduce (Reader-Retriever-QA-Space and Reader-Retriever-\{Q\}A-Space). The column ``Proposed vs SOTA'' indicates which rows to look at for comparing our method with state-of-the-art OpenQA methods, while the column ``Final vs Components'' indicates which rows to look at when examining the performance of individual workflows that are components of our final model \model.

The numbers in Table \ref{tab:accuracy_overall} indicate that the OpenQA task is hard, which is consistent with our discussion in Section~\ref{sec:related}. In particular, the DrQA test accuracy is lower than the author-reported number (27.1\%) on the Wikipedia dataset. This is not very surprising, because although both the DrQA paper and this paper use different datasets for training and testing, the similarity between the training and testing datasets is much higher in the DrQA paper than it is here. In the DrQA paper, the model is trained on SQuAD, which is based on 536 Wikipedia articles, and the test is done on the entire Wikipedia corpus with the questions selected from SQuAD. This means the structure and format of the questions and documents for training and testing can be similar for DrQA. 
%. However, in this paper the similarity between the training and testing datasets lies in the facts that the SQuAD training corpus is a subset of Wikipedia, the TriviaQA-Open testing corpus contains another subset of Wikipedia articles, and more than 90\% of the true answer entities in TriviaQA-Open are Wikipedia titles. Such similarity is much weaker than that in the DrQA paper, because TriviaQA-Open also has a large body of web articles that look very different from Wikipedia ones, and the Wikipedia page titles do not contribute anything during the training. 
That may explain why DrQA is reported to work better by its authors.

Despite the difficulty of the OpenQA task and possible dissimilarity between the training and testing datasets, our proposed method \model~outperforms both DrQA and BERT by a very big margin. This margin is about six times larger than the difference in performance between DrQA and BERT. If the 2\% difference between DrQA and BERT represents the consequence of differences in the detailed design of the retriever and reader modules in a retriever-reader model (e.g. TF-IDF vs semantic embedding, RNN vs BERT), then the 12\% margin between \model~and DrQA should be largely credited to the essential differences in the overall model structures.

When individual components of \model~are inspected, our novel reader-retriever model structure alone on the \{Q\}A Space also outperforms DrQA and BERT, with a smaller margin though. Our reader-retriever component on the QA Space is not working well by itself, but as an integral part of the answer aggregation mechanism, it helps push up the performance of our final model \model. The final model \model~performs significantly better than any of its component models, showing the importance of its integrity.

\begin{table}[ht]
  \centering
  \caption{Test accuracy on TriviaQA-Open.}
  \label{tab:accuracy_overall}
  \begin{tabular}{cccc}
    \toprule
    Method & Accuracy & Proposed vs SOTA & Final vs Components\\
    \midrule
    \model & \textbf{0.30} & $\bullet$ & $\bullet$ \\
    DrQA & 0.18 & $\bullet$ & \\
    Retriever-Reader-BERT-Large & 0.16 & $\bullet$ & $\bullet$\\
    Retriever-Reader-BERT-Base & 0.15 & $\bullet$ & $\bullet$\\
    Reader-Retriever-QA-Space & 0.07 & & $\bullet$\\
    Reader-Retriever-\{Q\}A-Space & 0.21 & & $\bullet$\\
  \bottomrule
\end{tabular}
\end{table}

\subsection{Test Accuracy for Various Answer Types}

We further ask how the discussed algorithms work for different types of questions or answers. Following the same practice as in the TriviaQA paper \cite{joshi2017triviaqa}, we sample 200 question-answer pairs from TriviaQA-Open and manually analyze their properties. We find that about 36\% of those questions have person names or organization names as answers, 26\% ask for locations, and 38\% are expecting answers that are neither persons/organizations nor locations. This sample distribution is roughly consistent with what TriviaQA authors have reported (32\%, 23\%, and 45\% respectively). We then use this sampled dataset TriviaQA-Open-200 to evaluate the test accuracy of the methods for different types of questions and answers. We drop Retriever-Reader-BERT-Large for this analysis because its overall accuracy is very close to Retriever-Reader-BERT-Base (Table \ref{tab:accuracy_overall}) but it consumes much more computational resources.

The results of this experiment are shown in Table \ref{tab:accuracy_breakdown}. Among the three types, questions that ask for Person/Organization names or locations look significantly easier to answer than those asking for other miscellaneous things for all algorithms, and our proposed method \model~takes the lead. Among the other models, it looks like BERT is good at questions about Person/Organization names and our newly proposed reader-retriever algorithm on the \{Q\}A Space is good at answering questions for locations. On the other hand, when the expected answer is neither a person/organization nor a location, DrQA still has some chance of getting the right answer, while all other methods including ours almost always fail. This is probably due to the fact that our methods rely on NER (Figure \ref{fig:model-structure}) but DrQA does not. How to better answer those miscellaneous questions remains an interesting future direction.

\begin{table}[ht]
  \centering
  \caption{Test accuracy on TriviaQA-Open-200 for various answer types.}
  \label{tab:accuracy_breakdown}
  \begin{tabular}{ccccc}
    \toprule
    Method & Overall & Person/Org (36\%) & Location (26\%) & Others (38\%)\\
    \midrule
    \model & \textbf{0.34} & \textbf{0.56} & \textbf{0.46} & 0.05\\
    DrQA & 0.22 & 0.33 & 0.23 & \textbf{0.11}\\
    Retriever-Reader-BERT-Base & 0.20 & 0.39 & 0.23 & 0\\
    Reader-Retriever-QA-Space & 0.10 & 0.22 & 0.08 & 0\\
    Reader-Retriever-\{Q\}A-Space & 0.22 & 0.28 & 0.38 & 0.05\\
  \bottomrule
\end{tabular}
\end{table}

\subsection{Notes on Question Space Quality}

A manual inspection into the constructed question spaces reveals three aspects of it that are worth discussion. 1) Many questions look reasonable, and those example questions shown in Figure \ref{fig:1} are actually real examples taken from our \{Q\}A Space that are associated with the answer ``Chicago Bears''. 2) There are also many questions that to some extent deviate from being a ``correct'' question to ask for a given answer. One frequently observed mistake is the use of a wrong question word. 3) Some questions like ``who did Bob talk to'' are reasonable and answerable given the context, but does not make sense when asked in general. Since \model~relies on the generated questions, its performance is hopeful to get further enhanced if the quality in the question spaces can be improved. How to do better question generation for OpenQA is left for future work.

\section{Case Study: Why It Works?}

Now let's refer back to the example in Figure \ref{fig:1} to get more intuition about why question spaces work. The fact is that the question ``who won Super Bowl XX'' is a real question from TriviaQA, the right branch of Figure \ref{fig:1} is the only real result observed in experiments, and the left and middle branches show some ideal results that do not actually happen. In reality, the left branch retrieves an irrelevant passage and thus gets a wrong answer, while the middle branch also fails because our entity recognition module fails to recognize the entity ``Super Bowl XX''. On the other hand, it does recognizes the entity ``Chicago Bears'', and the question space does contain clues that connect the correct answer to the query. Although there is no question in the QA Space that exactly matches the given query, by aggregating the signal from multiple questions in the \{Q\}A Space, our algorithm succeeds in finding the answer.

\section{Conclusion}

We propose \model, a novel algorithm that constructs question spaces from corpora and uses them to improve OpenQA performance. While many state-of-the-art methods follow a design called the retriever-reader, one key component of \model~is a novel OpenQA model structure called the reader-retriever that is different from the design of any existing work. Experiments on public datasets show that our proposed method outperforms two retriever-reader baselines by a large margin that is much greater than the difference between the two baselines. A more detailed examination of the results reveals that our method has the potential to get further improved if solutions can be proposed in future work to handle less typical question types better or generate more realistic questions.

\bibliographystyle{acm}
\bibliography{main}

\end{document}